\documentclass{article}

\usepackage[nonatbib,final]{neurips_2024}

\usepackage[utf8]{inputenc} 
\usepackage[T1]{fontenc}    
\usepackage{hyperref}       
\usepackage{url}            
\usepackage{booktabs}       
\usepackage{amsfonts}       
\usepackage{nicefrac}       
\usepackage{microtype}      
\usepackage{xcolor}         

\usepackage{amssymb}
\usepackage{graphicx}
\usepackage{mathtools}
\usepackage{multirow}
\usepackage{color}
\usepackage{verbatim}
\usepackage{amsmath}
\usepackage{amssymb}
\usepackage{tablefootnote}
\usepackage{booktabs}
\usepackage{comment}
\usepackage{float}
\DeclareMathOperator*{\argmax}{arg\,max}

\usepackage{adjustbox}
\usepackage{graphicx}

\title{Is network fragmentation a useful complexity measure?}

\author{%
  Coenraad Mouton$^{1,2}$ \quad Randle Rabe$^{1,2}$ \quad Daniël G. Haasbroek$^{1,2, 3}$ \\
  \textbf{Marthinus W. Theunissen$^{1,2}$} \quad \textbf{Hermanus L. Potgieter$^{1,2,3}$} \quad \textbf{Marelie H. Davel$^{1,2,4}$} \\
  $^1$Faculty of Engineering, North-West University, South Africa \\ \quad $^2$Centre for Artificial Intelligence Research, South Africa \\ \quad $^3$South African National Space Agency \\ \quad $^4$National Institute for Theoretical and Computational Sciences, South Africa\\
  \texttt{\{moutoncoenraad, randlerabe\}@gmail.com} \\
}

\begin{document}

\maketitle

\begin{abstract}
It has been observed that the input space of deep neural network classifiers can exhibit `fragmentation', where the model function rapidly changes class as the input space is traversed.
The severity of this fragmentation tends to follow the double descent curve, achieving a maximum at the interpolation regime.
We study this phenomenon in the context of image classification and ask whether fragmentation could be predictive of generalization performance. 
Using a fragmentation-based complexity measure, we show this to be possible by achieving good performance on the PGDL (Predicting Generalization in Deep Learning) benchmark.
In addition, we report on new observations related to fragmentation, namely 
(i) fragmentation is not limited to the input space but occurs in the hidden representations as well, 
(ii) fragmentation follows the trends in the validation error throughout training, and
(iii) fragmentation is not a direct result of increased weight norms.
Together, this indicates that fragmentation is a phenomenon worth investigating further when studying the generalization ability of deep neural networks.
\end{abstract}

\section{Introduction}

Understanding and predicting the generalization ability of neural networks is an active area of research, both from an empirical and theoretical perspective.  Of particular interest is the empirically observed phenomenon of `double descent'. Double descent (DD) refers to the observation that when measuring generalization error as a function of increasing capacity, deep neural networks (DNNs) do not exhibit the U-curve typical of classical machine learning but can, in addition, produce a second descent where generalization error continues to decrease as model capacity increases beyond a critical point~\cite{Belkin2019ReconcilingMM}. 

Various studies investigate DD and the underlying mechanisms by which it occurs~\cite{Nakkiran2020DeepDD, Spigler2018AJT,Ba2020Generalization, dAscoli2020DoubleTI}. Recently, Somepalli et al.~\cite{twice_dd_real_ver} observed that critically parameterized DNNs exhibit `fragmentation', which refers to the observation that a DNN's input space predictions can fragment into many different class regions. In addition, it was shown that the degree of fragmentation exhibited by a model is highly correlated with its test error when traversing the DD curve. 

Investigations of DD are closely related to studies that attempt to define a useful measure of `network complexity'~\cite{montufar_number_2014, raghu_expressive_2017, novak2018sensitivity, hanin_complexity_2019, gamba_are_2022}, meaning a measure that is capable of accurately measuring the effective, as opposed to representational, complexity of a DNN model's learned function. Complexity measures for DNNs are commonly investigated in the empirical context of predicting or ranking the generalization ability of a group of models by only relying on training data. 

In this work, we investigate both these areas (DD and complexity measures) from the perspective of network fragmentation. 
We show that not only does fragmentation correlate with test error in the controlled DD setting, but also in the more general setting of comparing models trained with different hyperparameter and architectural setups in the context of generalization prediction. Furthermore, we provide additional insights into the fragmentation phenomenon by studying its relation to hidden layers, the weight space, and dynamics during training. 


\section{Fragmentation and double descent}

We first confirm the observation by Somepalli et al. that input space fragmentation follows a DD~\cite{twice_dd_real_ver}, before ascertaining whether fragmentation can also be observed at the models' hidden layers.
%
\label{sec:dd_setup}
%
We describe how fragmentation is measured, develop a set of models to elicit DD, and define a straightforward process to investigate fragmentation in this context, before presenting results. 

\textbf{Measuring fragmentation} 
We measure fragmentation in the same way as Somepalli et al.~\cite{twice_dd_real_ver} but alter the definition to apply to any representational space within a DNN. Specifically, we sample $3$ random training samples (a triplet) and create a plane spanned by these samples. We then densely sample this plane and record the top-class predictions for each point within the plane. The fragmentation of the plane is measured by counting the number of distinct classification regions (sets of connected points with the same class prediction). When considering hidden layers, the process remains identical except that the plane is created by using the representations of the $3$ training samples found at a specific hidden layer in the network. 
See Appendix~\ref{sec:app_measuring_fragmentation} for additional detail.

\begin{figure}[H]
    \centering
    \includegraphics[width=0.98\linewidth]{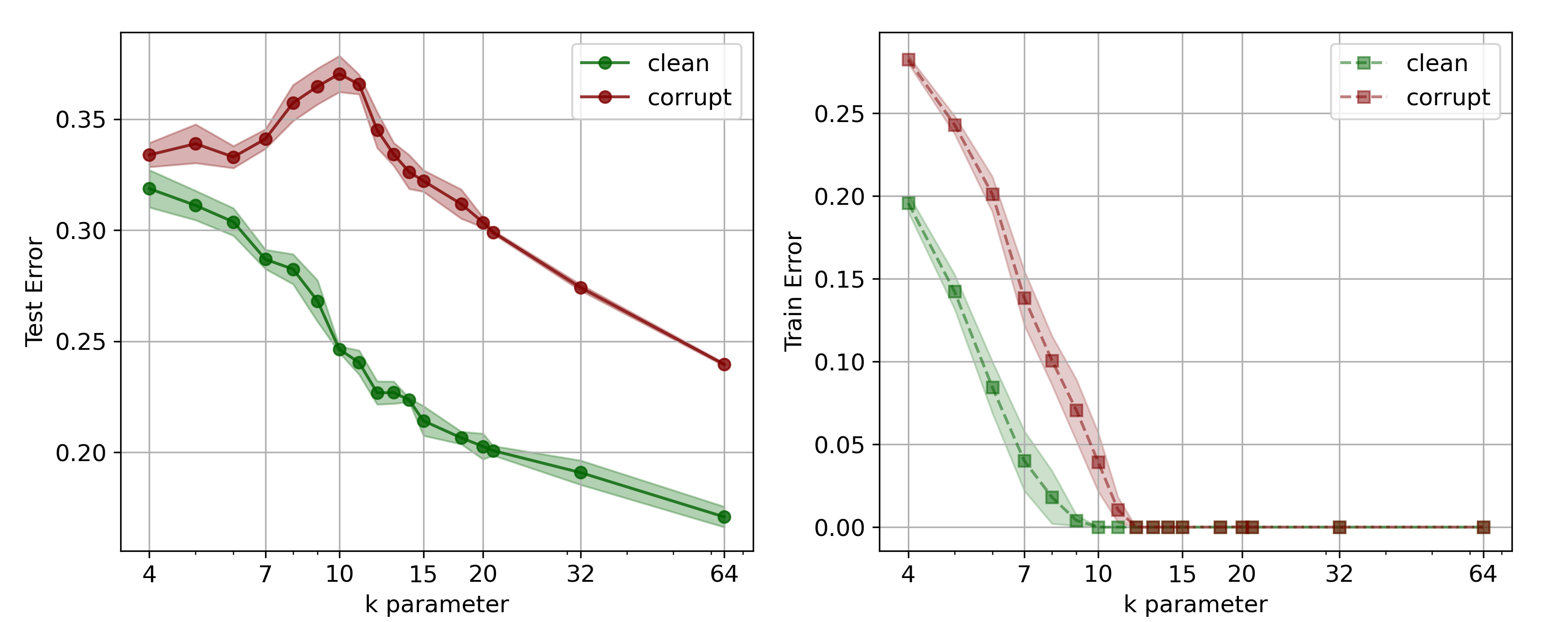}
    \caption{Error versus model capacity for models trained on partially label-corrupted data (red) and clean data (green). Left: Test error. Right: Train error.}
    \label{fig:input_and_hidden_dd_train_and_test_error}
\end{figure}

\textbf{DD models} 
We train two sets of CNN models with varying representational capacity on the CIFAR10 dataset: one set using the original data set, and the other using a label-corrupted version, in order to produce a pronounced DD. Specifically, $10\%$ of samples in the training set are randomly assigned a different label. We make use of an architecture similar to that of the `standard CNN' used by Nakkiran et al.~\cite{Nakkiran2020DeepDD}, which consists of $4$ convolutional layers with $[k, 2k, 4k, 8k]$ output channels, respectively. We choose values of $k$ between $4$ and $64$ to create a group of models with varying capacity. We train each set using $3$ different initialization seeds.  The train and test error for both sets of models are shown in Figure~\ref{fig:input_and_hidden_dd_train_and_test_error}. As expected, we observe a pronounced DD for the label-corrupted models.
See Appendix~\ref{sec:app_dd_details} for details on other training hyperparameters.

\textbf{Procedure}
For each model in each set, we measure the mean fragmentation as the average fragmentation over $500$ randomly sampled triplets. We construct the triplets by randomly sampling $3$ samples of the same class (without replacement). Furthermore, we use the same triplets for each model, and each triplet plane is sampled using $2\ 500$ equally spaced points. When measuring fragmentation at the hidden space, we measure the fragmentation at the output of each convolutional layer (post activation function) for each model.

\begin{figure}[H]
    \centering
    \includegraphics[width=0.49\linewidth]{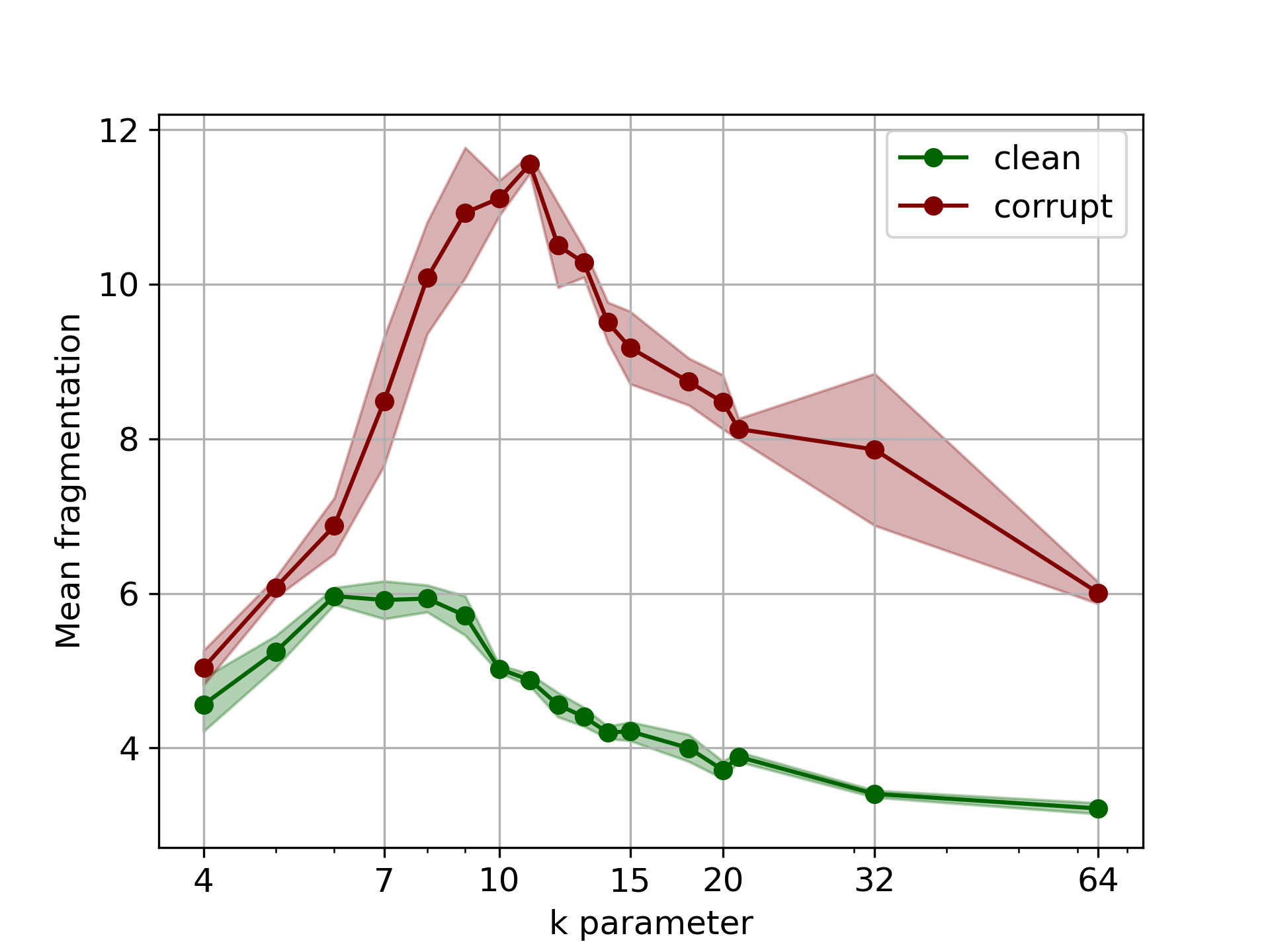}
    \includegraphics[width=0.49\linewidth]{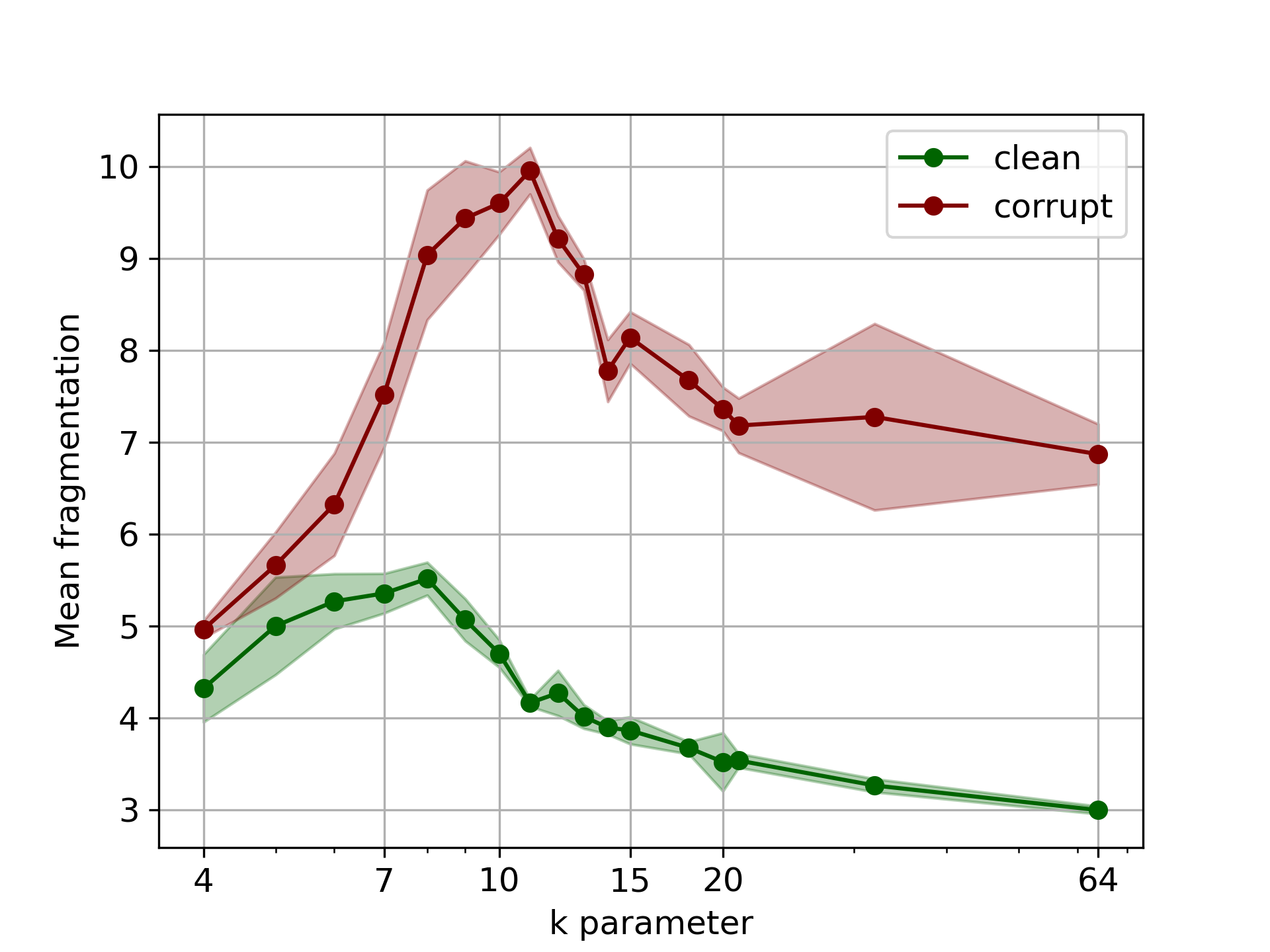}
    \caption{Mean fragmentation versus model capacity. Left: Input space. Right: Hidden space fragmentation for the first convolutional layer. `Clean' and `corrupt' refers to models trained on clean or partially label-corrupted data, respectively. Results are averaged over three seeds. The shaded regions indicate the error (standard deviation).}
    \label{fig:input_and_hidden_dd_fragmentation}
\end{figure}

\textbf{Results}
We show the mean fragmentation score per model (averaged over 3 seeds) in Figure~\ref{fig:input_and_hidden_dd_fragmentation}, which includes both the input space (left) and the hidden space of the first convolutional layer (right).
We observe that our input space results closely match those of Somepalli et al.; in particular, the mean fragmentation curve closely follows the test-error curve (see Figure~\ref{fig:input_and_hidden_dd_train_and_test_error} in Appendix~\ref{sec:app_dd_details}), and a clear second descent is visible. 
We also confirm that the mean fragmentation score is more pronounced for the corrupt than the clean case.
When we perform a similar analysis but measure the mean fragmentation of selected models during training (as opposed to after training) we find a similar correlation with validation error.
(See Figure~\ref{fig:training_metrics} in Appendix~\ref{sec:app_frag_during_training}.)

The fragmentation at the hidden space of the first convolutional layer shows similar trends to the input space, and we also find the same behavior at the other hidden layers, see Appendix~\ref{sec:app_hidden_depth}. Furthermore, we also observe that the mean fragmentation gradually decreases as function of depth, implying the fragmentation of the classification regions are unevenly distributed throughout the network.
%
These results confirm the observations of Somepalli et al., and extends their results to fragmentation behaviour during training, as well as to the networks' hidden representation space. 


\section{Fragmentation and generalization prediction}

In the previous section we have shown that fragmentation is highly correlated with test error in the controlled DD setting. We now turn our attention to the more general setting of determining the relationship between fragmentation and generalization performance for models with varying hyperparameters and architectural setups. Specifically, we rely on the Predicting Generalization in Deep Learning (PGDL)~\cite{pgdl_overview} benchmark.

The PGDL benchmark is a `dataset' of trained CNN image classification models split into $8$ different tasks. Each task refers to a set of models of a certain architectural family trained on the same dataset. The goal of this benchmark is to develop some measurement of the model and its training data that is able to rank the models within each task according to their generalization gap, i.e. a complexity measure. The complexity  measure is then evaluated by determining how well this ranking aligns with the true generalization gap ranking. 

The tasks within the dataset are split into two groups. Tasks $1$ to $5$ form the development set, with the intended usage of developing and refining a complexity measure. Tasks $6$ to $9$ forms the test set, which should be used as a held out set to evaluate the complexity measure, i.e. these tasks should only be evaluated once, to prevent overfitting.  


\subsection{Refined metrics}

In addition to the original fragmentation metric, we define two variations that take into account other characteristics of the classification regions.  Where the original metric considers the number of regions, these metrics use the relative sizes of the regions.  Specifically, we hypothesize that generalization performance is negatively correlated with larger `foreign' regions, and, therefore, we consider the fraction of the sampled-region area covered by these `foreign' classification regions.

We define \emph{foreign-region coverage} as the combined area of all regions that do not contain any of the triplet points, divided by the total area of the sampled region.  Similarly, we define \emph{foreign-class coverage} as the combined area of any regions that do not have the same class as any of the triplet points, divided by the total area.  These metrics allow us to explore the correlation between generalization and the relative sizes of the `foreign' regions.

\subsection{Predicting generalization} 
\label{sec:gen_prediction}
We calculate the mean fragmentation, foreign-region coverage, and foreign-class coverage for each model and task in the PGDL dataset. We rely on the same settings as used earlier in Section~\ref{sec:dd_setup}.
%
To assess the relationship between each complexity measure and the true generalization gap of the models, we calculate the conditional mutual information (CMI) between the two, conditioned on each set of hyperparameters\footnote{This metric measures the minimal conditional mutual information across all sets of hyperparameter combinations. See Jiang et al.~\cite{pgdl_overview,fantastic_gen} for a comprehensive explanation.}, which is the standard evaluation metric for the PGDL benchmark~\cite{pgdl_overview}. In addition, we also compare our scores to three other complexity measures that currently have the highest average performance on the test set of tasks that we are aware of: DBI*LWM~\cite{rep_based_complexity_pgdl}, PCA Gi\&Mi~\cite{schiff2021predicting}, and Constrained Margin~\cite{input_can_predict_too}.

The CMI scores for our {\em input space} fragmentation metrics and those we compare with are shown in Table~\ref{tab:pgdl_cmi_results}. Note that we present the mean scores for the PGDL development set (Tasks $1$ to $5$) and test set (Tasks $6$ to $9$) separately. 
\begin{table}[h]
\centering
\caption{Conditional Mutual Information (CMI) score for several complexity measures on the PGDL benchmark. `FRC' refers to foreign-region coverage, and `FCC' refers to foreign-class coverage. `Dev mean' is calculated as the average score over Tasks $1$ to $5$, and `Test mean' over Tasks $6$ to $9$. There is no Task $3$. The top two solutions for each task are highlighted in bold. }
\label{tab:pgdl_cmi_results}
\begin{adjustbox}{max width=\textwidth}
\begin{tabular}{@{}ccccccc@{}}
\toprule
\multirow{2}{*}{Task} & Natekar and Sharma~\cite{rep_based_complexity_pgdl} & Schiff et al.~\cite{schiff2021predicting} & Mouton et al.~\cite{input_can_predict_too} & \multicolumn{3}{c}{Fragmentation metrics} \\
 & DBI*LWM & PCA Gi\&Mi & Constrained Margin & Fragmentation & FRC & FCC \\ \midrule
1 & 00.00 & 00.04 & \textbf{39.49} & \textbf{03.25} & 00.08 & 00.08 \\
2 & 32.05 & \textbf{38.08} & \textbf{51.98} & 13.09 & 26.16 & 27.64 \\
4 & \textbf{31.79} & \textbf{33.76} & 21.44 & 15.92 & 29.31 & 29.49 \\
5 & 15.92 & 20.33 & 04.93 & \textbf{31.23} & \textbf{23.41} & 22.61 \\ \midrule
Dev mean & 19.94 & \textbf{23.05} & \textbf{29.46} & 15.87 & 19.74 & 19.96 \\ \midrule
6 & \textbf{43.99} & 40.06 & 30.83 & 37.14 & \textbf{40.32} & 40.10 \\
7 & 12.59 & 13.19 & 13.26 & 02.53 & \textbf{16.45} & \textbf{16.45} \\
8 & 09.24 & 10.30 & \textbf{13.48} & \textbf{10.91} & 09.84 & 09.99 \\
9 & 25.86 & 33.16 & \textbf{51.46} & 17.46 & 37.00 & \textbf{37.24} \\ \midrule
Test mean & 22.92 & 24.18 & \textbf{27.26} & 17.01 & 25.90 & \textbf{25.94} \\ \bottomrule
\end{tabular}
\end{adjustbox}
\end{table}

We observe that fragmentation and the two additional fragmentation-based metrics achieve competitive scores. This shows that an increase in fragmentation is correlated with a decrease in generalization performance. Interestingly, we find that foreign-region coverage and foreign-class coverage outperform the base fragmentation score on average, which points to the notion that considering the size of the classification regions is beneficial. Furthermore, we note that foreign-class coverage achieves the second-highest mean test set performance, only trailing behind the recently proposed constrained margin complexity measure~\cite{input_can_predict_too}. 
We find that input space fragmentation is more predictive of generalization than hidden space fragmentation, a metric we investigate in Appendix~\ref{sec:app_hidden_gen_prediction}.

\section{Discussion and conclusion}

We have shown that fragmentation can be a useful complexity measure of deep neural networks. Not only does fragmentation correlate well with test error in the DD setting, fragmentation-based metrics can be utilised as a measure for generalization prediction.  

While we have shown the utility of fragmentation as a complexity measure, the underlying mechanisms causing classification regions to fragment remains unknown. We believe that fragmentation is closely related to the smoothness of the learned function.
In fact, Somepalli et al. hypothesize that at the point of interpolation on the DD curve, 
there are oscillations that result in poor output function behaviour which causes the observed instability around class boundaries and that these oscillations are not needed for interpolation. 

There are several other related works that study the smoothness of the input-output mapping in DNNs. Gamba et al.~\cite{gamba_are_2022} propose a local measure of model complexity (nonlinearity) called \emph{absolute deviation}. They showed that, at the point of interpolation, the model output function was measured to be at its most `complex' or nonlinear, resulting in comparatively less smooth functions and that the smoothness of the functions increased as model capacity moved into the overparameterized regime. Recently, it was demonstrated by Teney et al.~\cite{teney2024neural_redshift} that ReLU networks at initialization have an implicit bias towards smoothness and that this bias is independent of the model width, depth, or weight magnitude. Similarly, we note that fragmentation is very low at model initialization (see Figure~\ref{fig:training_metrics}).

We believe that understanding the underlying weight space mechanisms that cause the instability in the classification regions (or alternatively, the reduction in smoothness) is a logical next step. We have made some preliminary strides towards this goal, such as determining that the causes of fragmentation are distributed throughout the network (see Appendix~\ref{sec:app_hidden_depth}). 
We have also found that the underlying causes are not trivial, such as due to exploding weight norms (see Appendix~\ref{sec:app_frag_and_weight_norms}). 


In conclusion, we have investigated fragmentation as a complexity measure. 
%
%
We find that it shows promise in measuring DNN complexity. In addition, we hypothesize that a deeper understanding of the underlying mechanisms that cause fragmentation may shed light on more general questions related to the change in smoothness of the output function as capacity changes.

\begin{ack}

This work is based on research supported in part by the National Research Foundation of South Africa (Ref Numbers \textit{PSTD23042296065}, \textit{PSTD23042898868}, \textit{RA211019646111}).

\end{ack}
\bibliographystyle{unsrt}  
\bibliography{neurips_2024}

\begin{thebibliography}{10}

\bibitem{Belkin2019ReconcilingMM}
Mikhail Belkin, Daniel Hsu, Siyuan Ma, and Soumik Mandal.
\newblock Reconciling modern machine-learning practice and the classical bias–variance trade-off.
\newblock {\em Proceedings of the National Academy of Sciences}, 116(32):15849–15854, 2019.

\bibitem{Nakkiran2020DeepDD}
Preetum Nakkiran, Gal Kaplun, Yamini Bansal, Tristan Yang, Boaz Barak, and Ilya Sutskever.
\newblock Deep double descent: Where bigger models and more data hurt.
\newblock In {\em International Conference on Learning Representations}, 2020.

\bibitem{Spigler2018AJT}
Stefano Spigler, Mario Geiger, Stéphane d'Ascoli, Levent Sagun, Giulio Biroli, and Matthieu Wyart.
\newblock A jamming transition from under- to over-parametrization affects loss landscape and generalization.
\newblock {\em Journal of Physics A: Mathematical and Theoretical}, 52(47):474001, 2019.

\bibitem{Ba2020Generalization}
Jimmy Ba, Murat Erdogdu, Taiji Suzuki, Denny Wu, and Tianzong Zhang.
\newblock Generalization of two-layer neural networks: An asymptotic viewpoint.
\newblock In {\em International Conference on Learning Representations}, 2020.

\bibitem{dAscoli2020DoubleTI}
Stéphane D’Ascoli, Maria Refinetti, Giulio Biroli, and Florent Krzakala.
\newblock Double {trouble} in {double} {descent}: {Bias} and {variance}(s) in the {lazy} {regime}.
\newblock In {\em {International} {Conference} on {Machine} {Learning}}, 2020.

\bibitem{twice_dd_real_ver}
Gowthami Somepalli, Liam Fowl, Arpit Bansal, Ping Yeh-Chiang, Yehuda Dar, Richard Baraniuk, Micah Goldblum, and Tom Goldstein.
\newblock Can neural nets learn the same model twice? {I}nvestigating reproducibility and double descent from the decision boundary perspective.
\newblock In {\em IEEE/CVF Conference on Computer Vision and Pattern Recognition}, 2022.

\bibitem{montufar_number_2014}
Guido~F Montufar, Razvan Pascanu, Kyunghyun Cho, and Yoshua Bengio.
\newblock On the {number} of {linear} {regions} of {deep} {neural} {networks}.
\newblock In {\em Advances in {Neural} {Information} {Processing} {Systems}}, 2014.

\bibitem{raghu_expressive_2017}
Maithra Raghu, Ben Poole, Jon Kleinberg, Surya Ganguli, and Jascha Sohl-Dickstein.
\newblock On the {expressive} {power} of {deep} {neural} {networks}.
\newblock In {\em {International} {Conference} on {Machine} {Learning}}, 2017.

\bibitem{novak2018sensitivity}
Roman Novak, Yasaman Bahri, Daniel~A. Abolafia, Jeffrey Pennington, and Jascha Sohl-Dickstein.
\newblock Sensitivity and generalization in neural networks: an empirical study.
\newblock In {\em International Conference on Learning Representations}, 2018.

\bibitem{hanin_complexity_2019}
Boris Hanin and David Rolnick.
\newblock Complexity of {linear} {regions} in {deep} {networks}.
\newblock In {\em {International} {Conference} on {Machine} {Learning}}, 2019.

\bibitem{gamba_are_2022}
Matteo Gamba, Adrian Chmielewski-Anders, Josephine Sullivan, Hossein Azizpour, and Marten Bjorkman.
\newblock Are {all} {linear} {regions} {created} {equal}?
\newblock In {\em {International} {Conference} on {Artificial} {Intelligence} and {Statistics}}, 2022.

\bibitem{pgdl_overview}
Yiding Jiang, Pierre Foret, Scott Yak, Daniel~M Roy, Hossein Mobahi, Gintare~Karolina Dziugaite, Samy Bengio, Suriya Gunasekar, Isabelle Guyon, and Behnam Neyshabur.
\newblock {NeurIPS} 2020 competition: predicting generalization in deep learning.
\newblock {\em arXiv:2012.07976}, 2020.

\bibitem{fantastic_gen}
Yiding Jiang, Behnam Neyshabur, Hossein Mobahi, Dilip Krishnan, and Samy Bengio.
\newblock Fantastic generalization measures and where to find them.
\newblock In {\em International Conference on Learning Representations}, 2019.

\bibitem{rep_based_complexity_pgdl}
Parth Natekar and Manik Sharma.
\newblock {R}epresentation based complexity measures for predicting generalization in deep learning.
\newblock {\em arXiv:2012.02775}, 2020.

\bibitem{schiff2021predicting}
Yair Schiff, Brian Quanz, Payel Das, and Pin-Yu Chen.
\newblock Predicting deep neural network generalization with perturbation response curves.
\newblock In {\em {Advances in Neural Information Processing Systems}}, 2021.

\bibitem{input_can_predict_too}
Coenraad Mouton, Marthinus~Wilhelmus Theunissen, and Marelie~H. Davel.
\newblock Input {margins} {can} {predict} {generalization} {too}.
\newblock In {\em AAAI Conference on Artificial Intelligence}, 2024.

\bibitem{teney2024neural_redshift}
Damien Teney, Armand Nicolicioiu, Valentin Hartmann, and Ehsan Abbasnejad.
\newblock Neural redshift: Random networks are not random functions.
\newblock In {\em The IEEE/CVF Conference on Computer Vision and Pattern Recognition}, 2024.

\bibitem{kendall_corr}
Maurice~G Kendall.
\newblock {{A new measure of rank correlation}}.
\newblock {\em Biometrika}, 30(1-2):81--93, 06 1938.

\bibitem{distance_from_init}
Vaishnavh Nagarajan and J~Zico Kolter.
\newblock Generalization in deep networks: The role of distance from initialization.
\newblock {\em arXiv preprint arXiv:1901.01672}, 2019.

\end{thebibliography}

\newpage
\appendix
\section{Experimental details}

\subsection{Triplet Plane and Fragmentation}
\label{sec:app_measuring_fragmentation}

We measure fragmentation in the same way as \cite{twice_dd_real_ver}, although we alter some definitions to allow for measuring the fragmentation at any representational space within a DNN.

Consider an image dataset $\mathcal{D} = \{\mathbf{x}_i, y_i\}_{i=1}^{r}$ consisting of $r$ sample pairs where $\mathbf{x}_i \in \mathbb{R}^n$ is the n-dimensional input and $y_i \in \{0, 1, ..., k-1\}$ the label. Furthermore, we let $f_\theta:\ \mathbb{R}^n \rightarrow \mathbb{R}^{k}$ denote a family of neural networks with $L$ hidden layers and a set of trainable parameters $\theta$, where $\hat{y}_i = \argmax{f_{\theta}(\mathbf{x}_i)}$ is the model's prediction.  Finally, let $\mathbf{x}^\ell$, $\ell \in [0,L]$, correspond to the representation of a sample at a layer $\ell$, where $\mathbf{x}^{0} = \mathbf{x}$ denotes the input sample. For $\ell > 0$, $\mathbf{x}^{\ell}$ is the vector formed \emph{after} ReLU activation is applied element-wise.

For any point $\mathbf{x}^{\ell}$ in the latent space of layer $\ell > 0$, the classification is given by $\argmax{f_{\theta^{\ell+1}}(\mathbf{x}^\ell)}$, where $f_{\theta^{\ell+1}}$ is defined as the neural network with layers $0$ to $\ell$ removed from $f_{\theta}$. 

\paragraph{Triplet Plane} To construct a `triplet plane',  we randomly sample a triplet set of unique images $\{\mathbf{x}_1, \mathbf{x}_2, \mathbf{x}_3\}$ from $\mathcal{D}$ conditioned on $y_1 = y_2 = y_3$. Then, for a given layer $\ell$, we define the vectors $\mathbf{v}_1^{\ell}$ and $\mathbf{v}_2^\ell$ as $\mathbf{v}_1^{\ell} = \mathbf{x}_2^{\ell} - \mathbf{x}_1^{\ell}$ and $\mathbf{v}_2^\ell = \mathbf{x}_3^\ell - \mathbf{x}_1^\ell$. The plane $P = \text{span}_{\mathbb{R}}\{\hat{\mathbf{v}}_1^{\ell}, \hat{\mathbf{v}}_{1\bot}^{\ell} \}$ consists of all vectors spanned by the orthonormal basis vectors $\hat{\mathbf{v}}_1^{\ell}, \hat{\mathbf{v}}_{1\bot}^{\ell}$, where we have defined the orthogonal vector,
\begin{equation}
    \mathbf{v}_{1\bot}^{\ell} = \mathbf{v}_2^\ell - \text{proj}_{\mathbf{v}_1^\ell}\mathbf{v}_2^\ell, \quad \text{where proj}_{\mathbf{a}}\mathbf{b} = (\mathbf{b}\cdot \hat{\mathbf{a}})\ \hat{\mathbf{a}}.
\end{equation}
In practice, we sample points from the plane using the algorithm defined by \cite{twice_dd_real_ver}. More precisely, using $\mathbf{x}_{1}^{\ell}$ as the base image, we can find any point $\mathbf{s}^{\ell}$ on the plane $P$ with respect to the standard (feature) basis using the equation,
\begin{equation}
    \begin{aligned}
        \mathbf{s}^{\ell} &= \mathbf{x}_1^{\ell} + k_1 \hat{\mathbf{v}}_{1}^{\ell} + k_2 \hat{\mathbf{v}}_{1\bot}^{\ell}\\
        &= \mathbf{x}_1^{\ell} + (1 - \alpha)\ \text{min} \{ 0,\ \hat{\mathbf{v}}_1^{\ell} \cdot \mathbf{v}_2^{\ell}\}\ \hat{\mathbf{v}}_{1}^{\ell} + \alpha\ \text{max} \{ ||\mathbf{v}_1^{\ell}||,\ \hat{\mathbf{v}}_1^{\ell} \cdot \mathbf{v}_2^{\ell} \}\ \hat{\mathbf{v}}_{1}^{\ell} \\
        &\quad + \beta\ (\mathbf{v}_2^{\ell} \cdot  \hat{\mathbf{v}}_{1\bot}^{\ell})\ \hat{\mathbf{v}}_{1\bot}^{\ell},
    \end{aligned}
\end{equation}
where $k_1,\ k_2$ are bounded coordinates on the plane and $0 \leq \alpha,\ \beta \leq 1$. To understand the second equality, note that the range of the $k_2$ coordinate is between $0$ and the maximum value along $\hat{\mathbf{v}}_{1\bot}^{\ell}$, which is determined by the point $\mathbf{x}_3^{\ell}$. The minimum for the $k_1$ coordinate is dependent on the sign of $\hat{\mathbf{v}}_1^{\ell} \cdot \mathbf{v}_2^{\ell}$ (or, equivalently, the angle between the two vectors). Similarly, its maximum value is determined again by the sign of $\hat{\mathbf{v}}_1^{\ell} \cdot \mathbf{v}_2^{\ell}$ as well as the respective magnitudes $||\mathbf{v}_1^{\ell}||,\ |\hat{\mathbf{v}}_1^{\ell} \cdot \mathbf{v}_2^{\ell}|$. The second equality combines these ranges into a single equation parameterized by the variables $\alpha, \beta$. Finally, note that we can pad the plane by introducing a hyperparameter $\rho > 0$ and setting the parameter ranges to $-\rho \leq \alpha,\ \beta \leq 1 + \rho$.

\paragraph{Fragmentation} By sampling points from the triplet plane and passing the points through $f_{\theta^{\ell+1}}$, this results in a plane consisting of `classification regions': groups of points that are predicted as the same class. The fragmentation is then calculated by simply counting the number of distinct classification regions within the sampled plane.


\subsection{Double descent models}
\label{sec:app_dd_details}

Here we provide additional detail on the DD CNN models described in Section~\ref{sec:dd_setup}.
All models are trained on a $45\ 000/5\ 000$ train-validation split of the CIFAR10 dataset. As mentioned, we train one set on clean data, and another on label-corrupted data. For the label corruption, $10\%$ of the training labels are randomly assigned a different class. The exact same samples are label-corrupted for each model. Note that the validation and test set are never label-corrupted. 

Each model consists of $4$ convolutional layers, and one dense layer (excluding the output layer). The dense layer has a fixed size of $400$ nodes, while each CNN layer has $[k, 2k, 4k, 8k]$ channels, respectively. In terms of optimization, we rely on the Adam optimizer in combination with cross entropy loss. Each models is trained for $500$ epochs, without early stopping. We use a batch size of $256$ samples and an initial learning rate of $0.001$. We combine this with a step decay where the initial learning rate is multiplied by $0.99$ every $10$ epochs. This protocol ensures that all the models in the overparameterized regime ($k>9$ and $k > 11$ for clean and corrupt, respectively) fully interpolate the training data with $0\%$ train error. Refer back to Figure~\ref{fig:input_and_hidden_dd_train_and_test_error} for the relevant train and test error curves.



\section{Additional results}

\subsection{Hidden Space Fragmentation}
\label{sec:app_hidden_depth}

In this section, we provide additional details on fragmentation in the hidden layers.

\begin{figure}[tbh]
    \centering
    \includegraphics[width=0.49\linewidth]{Main/Figures/DD/cap_hidden_conv_1.png}
    \includegraphics[width=0.49\linewidth]{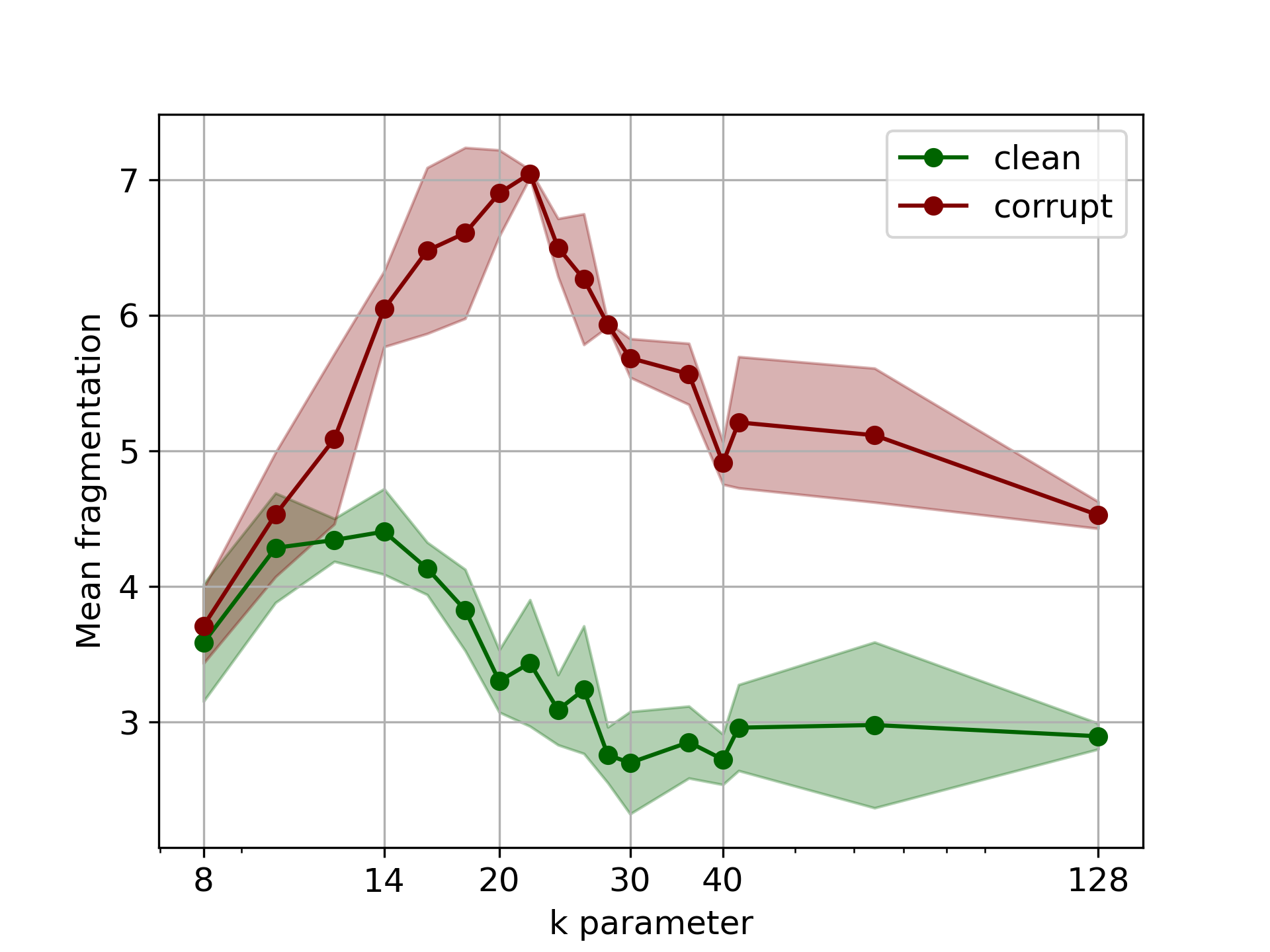}\\    \includegraphics[width=0.49\linewidth]{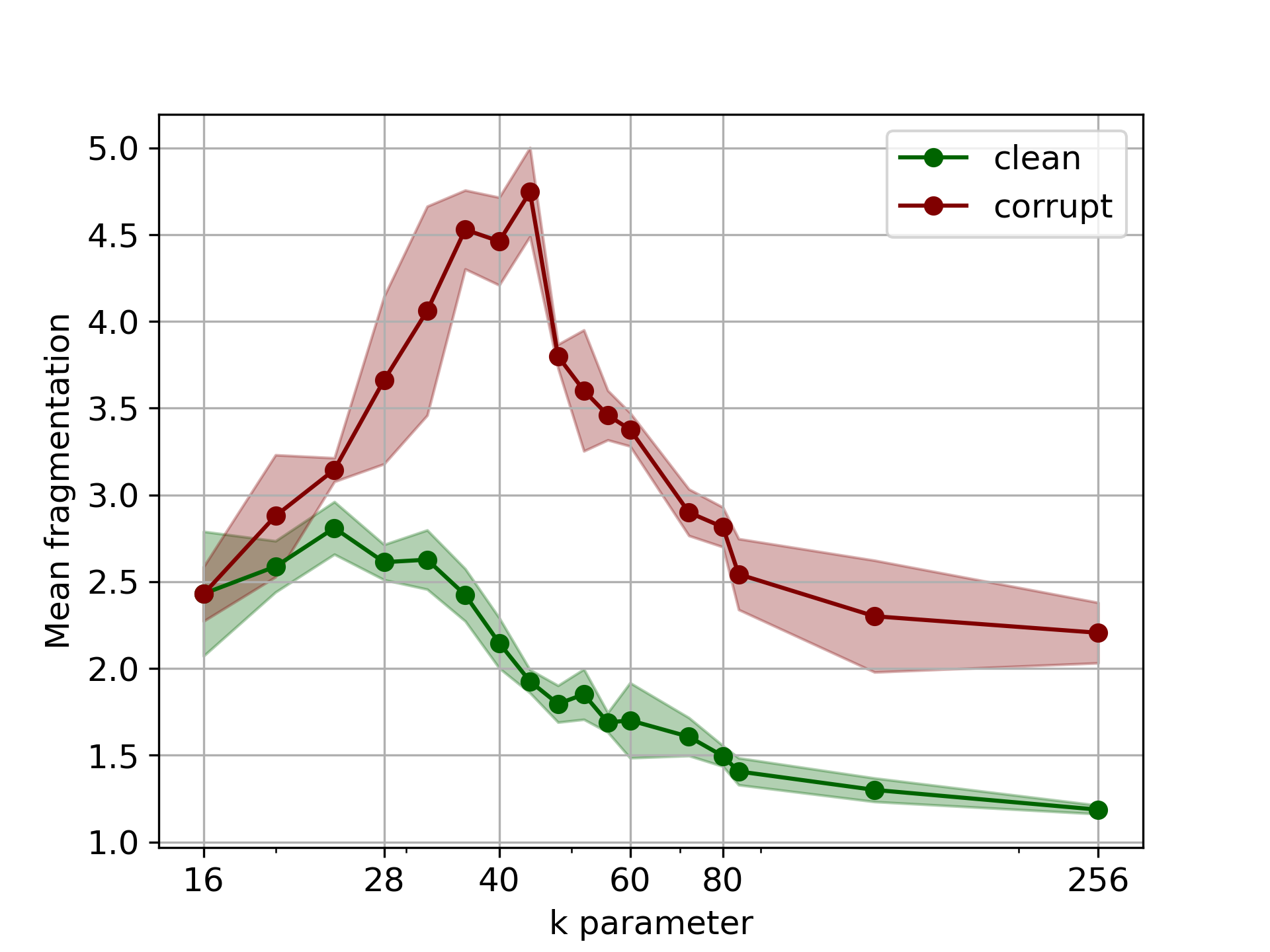}    \includegraphics[width=0.49\linewidth]{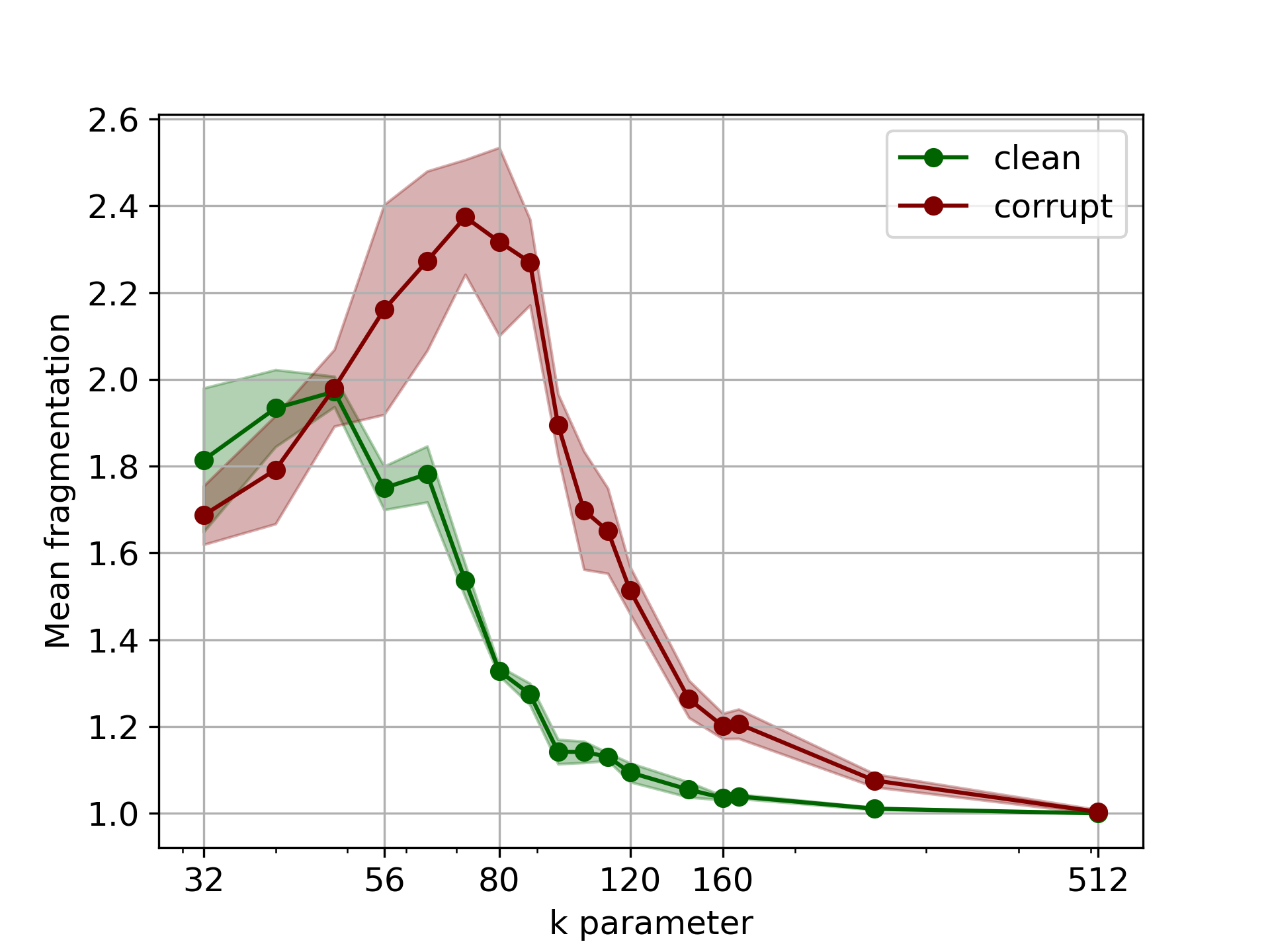}
    \caption{Mean fragmentation per convolutional layer versus model capacity. Channels per layer are given by the pattern $[k,\ 2k,\ 4k,\ 8k]$ where $k\in \{4,...,64\}$.  First Row: Fragmentation in the hidden space representations for convolutional layers $1$ and $2$, respectively. Second Row: Fragmentation in the hidden space representations for convolutional layers $3$ and $4$, respectively. Results are averaged over three seeds. The shaded regions indicate the error (standard deviation).}
    \label{fig:hidden_all_capacity_wise}
\end{figure}

Figure~\ref{fig:hidden_all_capacity_wise} shows fragmentation in each of the four hidden layers of the CNNs (recall that the hidden layers have channels $k,\ 2k,\ 4k,\ 8k$ where $k\in \{4,..., 64\}$). The top left plot shows how fragmentation changes as a function of capacity (measured by $k$) for the first hidden layer of all our CNNs. Likewise, the top right, bottom left, and bottom right figures show fragmentation for the remaining hidden layers.

\begin{figure}[H]
    \centering
    \includegraphics[width=0.49\linewidth]{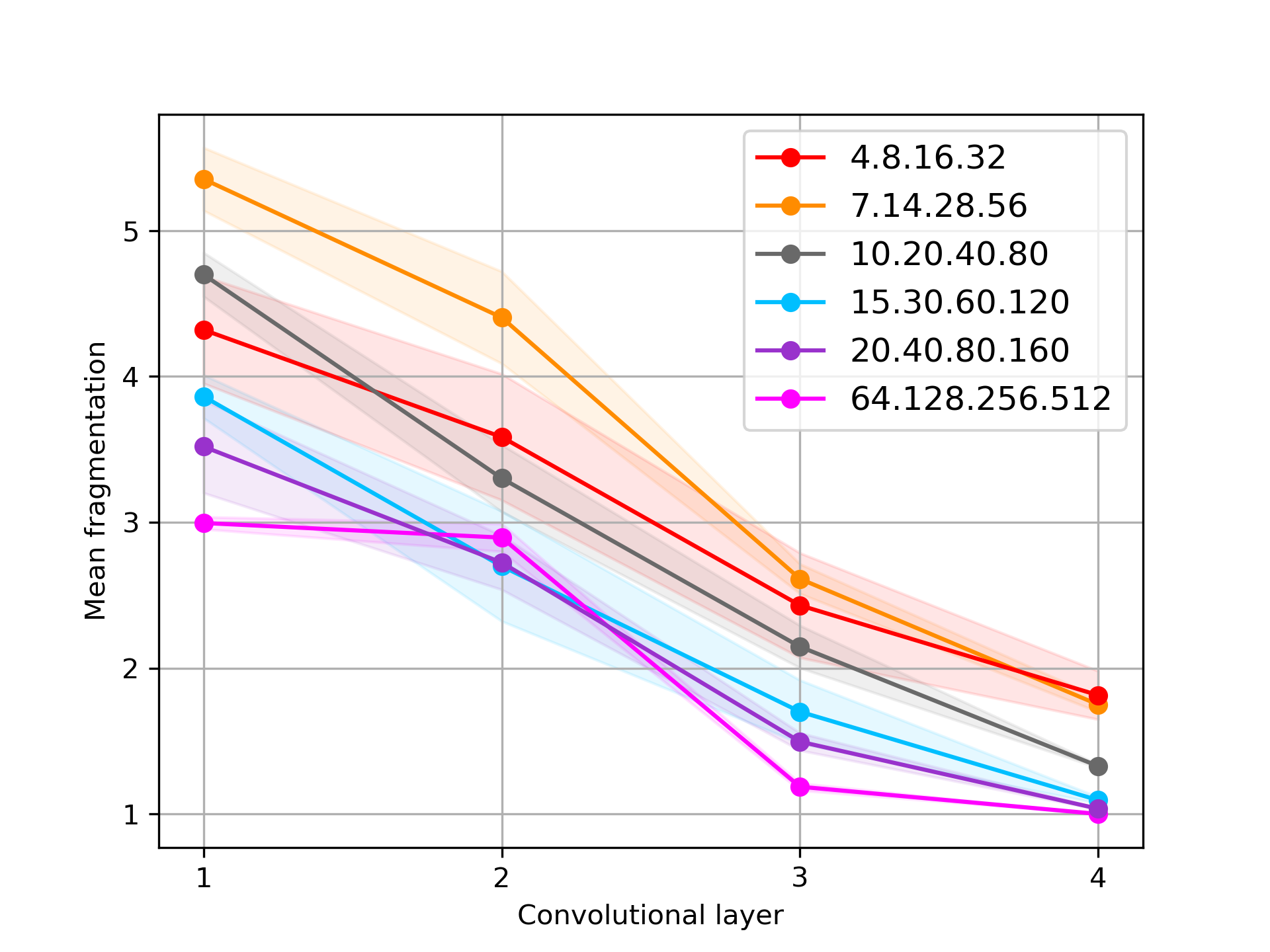}%
    \includegraphics[width=0.49\linewidth]{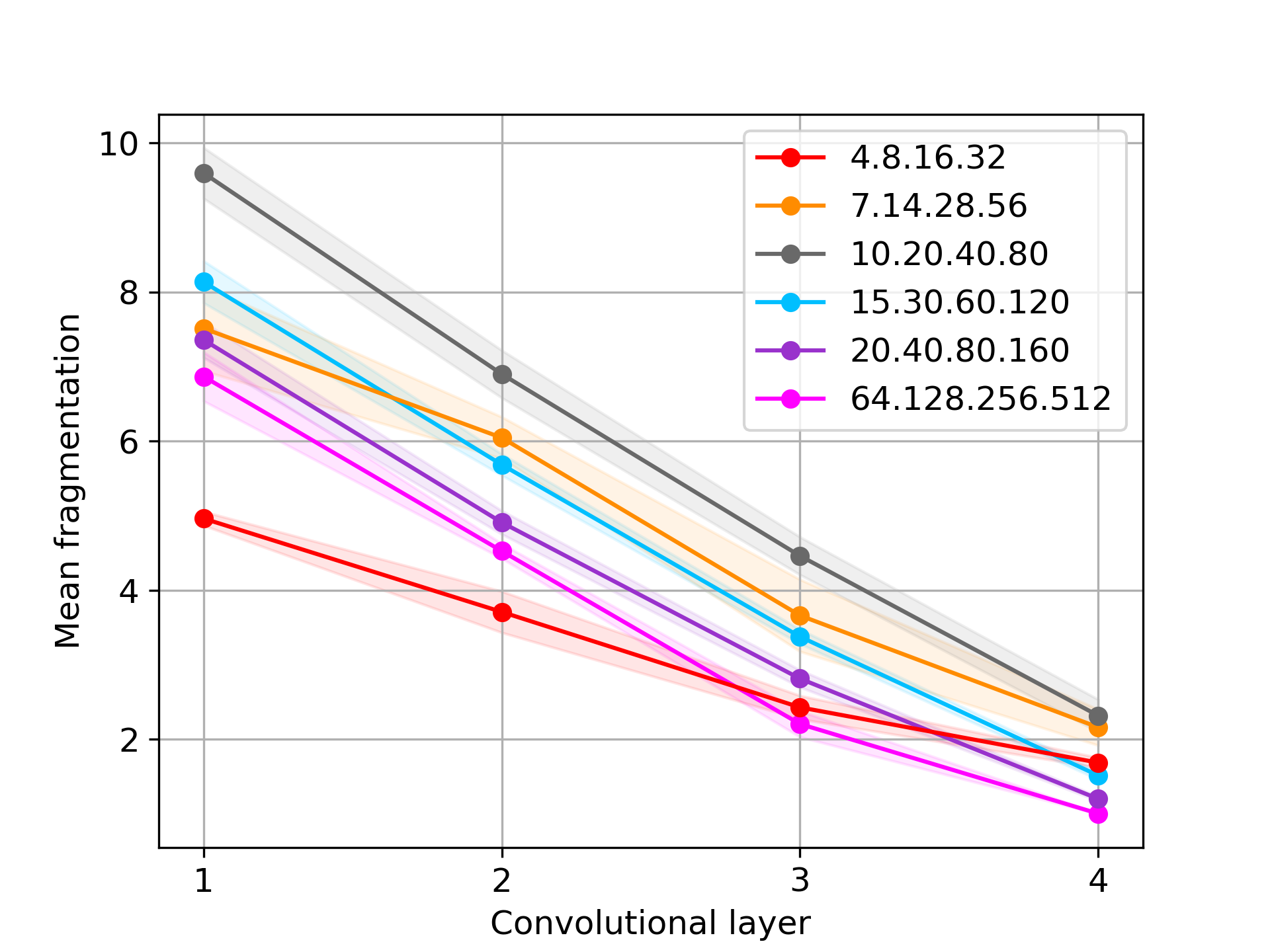}
    \caption{Mean fragmentation as a function of depth. Left: Depth-wise fragmentation on clean CIFAR10. Right: Depth-wise fragmentation on corrupted CIFAR10. Results are average over three seeds. The shaded regions indicate the error (standard deviation). The legend indicates the number of channels in each layer.}
    \label{fig:hidden_depthwise}
\end{figure}

Each figure exhibits a similar trend to the input space; in particular, fragmentation is observed to be more pronounced in the interpolation regime. Also, similar to the input space, we observe that the models trained on the corrupt data generally achieve higher mean fragmentation score relative to their clean counterparts. Finally, we also note that the mean fragmentation decreases as a function of depth, with the final hidden layer exhibiting the least fragmentation. This is made more apparent in Figure~\ref{fig:hidden_depthwise} which shows the depth-wise decrease in mean fragmentation for the clean case (left figure) and the corrupt case (right figure).

\subsection{Fragmentation during training}
\label{sec:app_frag_during_training}

We measure the fragmentation for a subset of the DD models (Sections~\ref{sec:dd_setup} and~\ref{sec:app_dd_details}) at several points throughout training.  For this we use the smallest underparameterized model ($k = 4$), the critically parameterized model ($k = 10$), and the largest over-parameterized model ($k = 64$).  The models are trained for $1\ 000$ epochs on both the original- and label-corrupted CIFAR10 dataset using the training setup described in Section~\ref{sec:app_dd_details}.

\begin{figure}[htb]
    \centering
    \includegraphics[width=\linewidth]{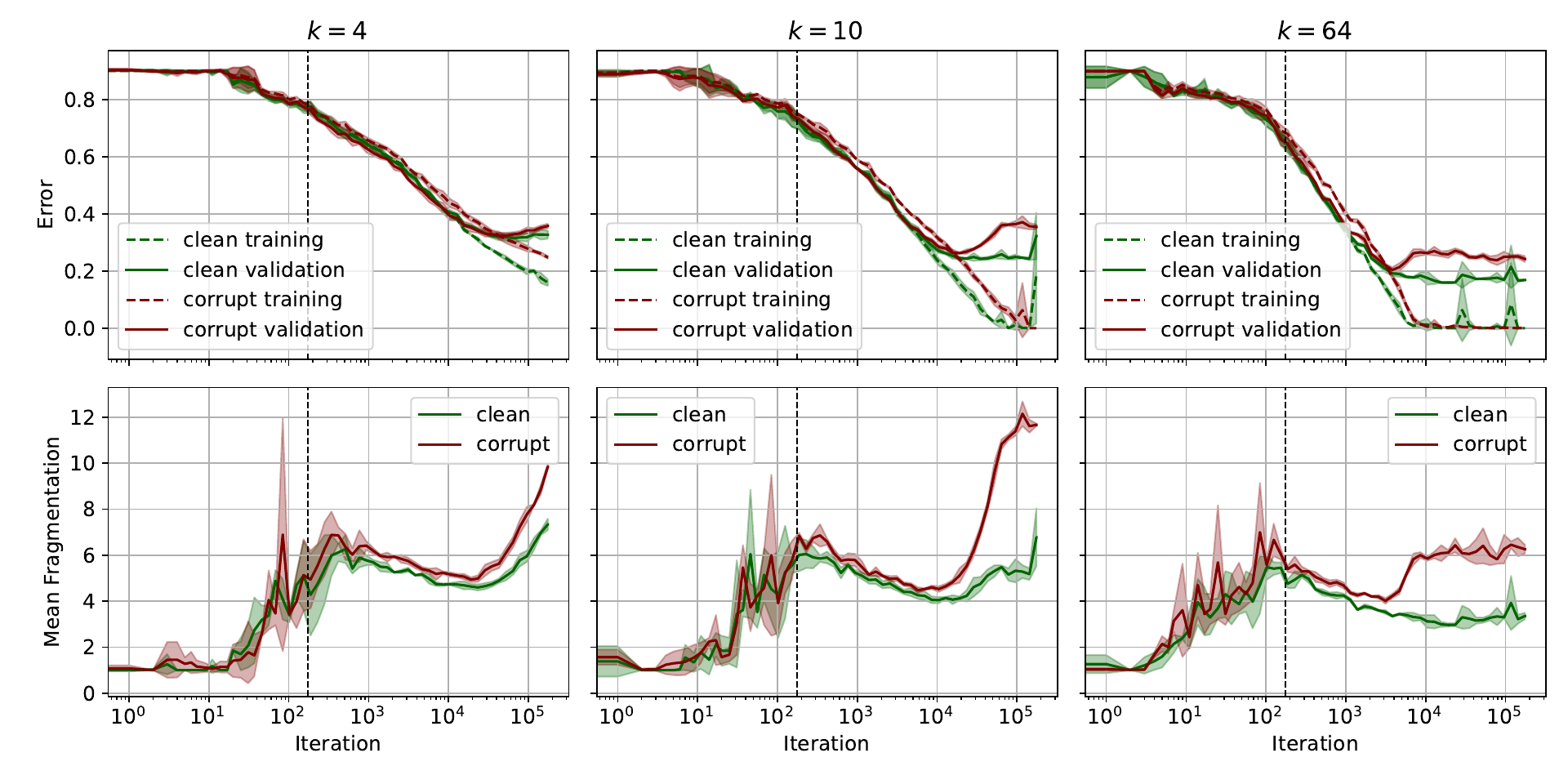}
    \caption{Mean fragmentation throughout training.  The dashed line marks the end of the first epoch.  Results are averaged over three seeds.  The shaded regions indicate the error (standard deviation).}
    \label{fig:training_metrics}
\end{figure}

The results in Figure~\ref{fig:training_metrics} show how fragmentation captures the divergence in validation performance between the clean and corrupt models, despite it being calculated on training data.  These results illustrate the correlation between fragmentation and performance on unseen data after the first epoch of training.

\subsection{Hidden layer generalization prediction}
\label{sec:app_hidden_gen_prediction}

In this section we compare the predictive performance of the fragmentation and foreign-class coverage metrics when measured at different layers in each model. Specifically, we consider the performance when measuring at the first and last hidden layer (separately). This provides us with an indication of how the performance of each metric varies as depth increases. 

Previously, in Section~\ref{sec:gen_prediction}, we used CMI to evaluate the performance of each measure. Here, we rather rely on a more interpretable metric, namely Kendall's rank correlation~\cite{kendall_corr}, which provides a score between $-1$ and $+1$. In this context, $+1$ indicates perfect agreement between the model ranking provided by the complexity measure and the ranking of the true generalization gap of the models. Conversely, $-1$ indicates perfect disagreement. The Kendall's rank correlation for each metric measured at each layer is shown in Table~\ref{tab:pgdl_kendall_input_and_hidden_compairson}.
\begin{table}[h]
\centering
\caption{Kendall's rank correlation for fragmentation and foreign-class coverage measured at different layers on the PGDL benchmark. `Dev mean' is calculated as the average score over Tasks $1$ to $5$, and `Test mean' over Tasks $6$ to $9$. There is no Task $3$. The best performing layer for each metric and task is highlighted in bold.}
\label{tab:pgdl_kendall_input_and_hidden_compairson}
\begin{tabular}{@{}cccc|ccc@{}}
\toprule
\multirow{2}{*}{Task} & \multicolumn{3}{c|}{Fragmentation} & \multicolumn{3}{c}{Foreign-class coverage} \\
 & Input space & First hidden & Last hidden & Input space & First hidden & Last hidden \\ \midrule
1 & -0.23 & -0.31 & \textbf{0.68} & 0.02 & -0.17 & \textbf{0.71} \\
2 & \textbf{0.76} & 0.62 & -0.29 & \textbf{0.84} & 0.74 & -0.26 \\
4 & \textbf{0.59} & 0.39 & 0.22 & \textbf{0.70} & 0.44 & 0.18 \\
5 & \textbf{0.66} & 0.61 & -0.05 & \textbf{0.59} & 0.50 & -0.03 \\ \midrule
Dev mean & \textbf{0.45} & 0.33 & 0.14 & \textbf{0.54} & 0.38 & 0.15 \\
\midrule
6 & \textbf{0.83} & 0.30 & 0.20 & \textbf{0.85 }& 0.33 & 0.22 \\
7 & \textbf{0.30} & -0.08 & 0.15 & \textbf{0.52} & 0.37 & 0.16 \\
8 & \textbf{0.53} & 0.37 & 0.33 & \textbf{0.52} & 0.38 & 0.32 \\
9 & \textbf{0.50} & 0.06 & 0.29 & \textbf{0.75} & 0.34 & 0.29 \\ \midrule
Test mean & \textbf{0.54} & 0.16 & 0.24 & \textbf{0.66} & 0.36 & 0.25 \\ \bottomrule
\end{tabular}
\end{table}

We observe that the input space offers the best performance by far. Furthermore, it is clear that the performance decreases from the first hidden layer to the last. This is to be expected, as the fragmentation greatly decreases as depth increases (see Appendix~\ref{sec:app_hidden_depth}). This implies that the measure likely becomes less discriminative, and therefore less useful, when measured at deeper points in the network. 

Peculiarly, we find that Task $1$ is an exception to the above, where the last hidden layer provides much better performance than the input space. 

\subsection{Fragmentation is not related to large weight norms}
\label{sec:app_frag_and_weight_norms}
Large weight norms are known to correlate with poor generalization, with weight norm regularization a standard element of the network optimization toolkit. 
We therefore ask whether fragmentation is not simply a side effect of large weight norms and evaluate parameter size for the same set of models described in Section \ref{sec:app_dd_details}.
Specifically, we measure both the Frobenius norm and the mean absolute value (mean L1 norm) of the tensors of parameters per layer, as representational capacity ($k$) is increased.
Using the Frobenius norm is more typical in this setting as it is indicative of the effect the weights has on the signal, but as this metric is sensitive to the difference in number of parameters per layer, we also report a mean value.

\begin{figure}[htb]
    \centering
    \includegraphics[width=0.49\linewidth]{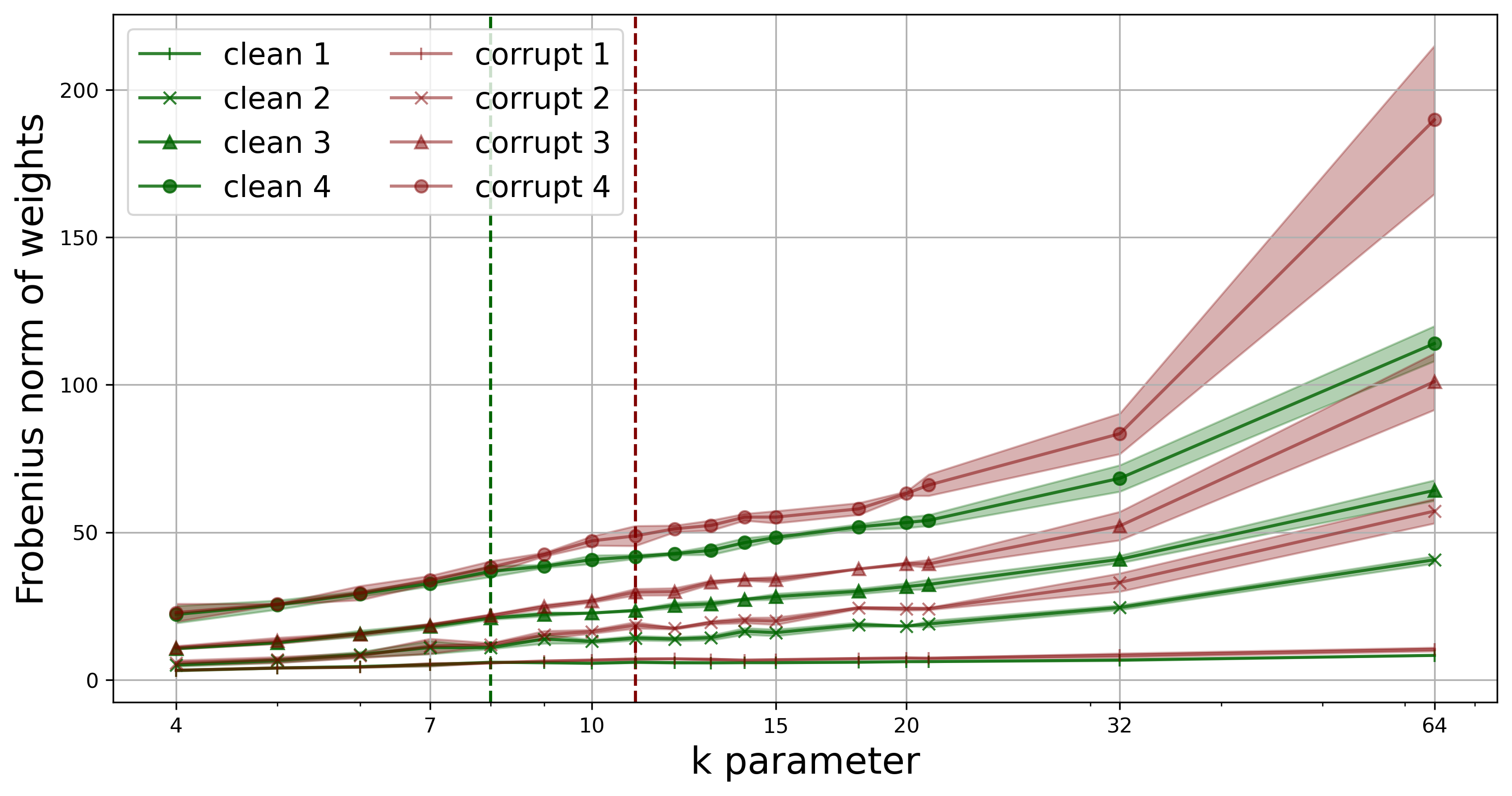}
    \includegraphics[width=0.49\linewidth]{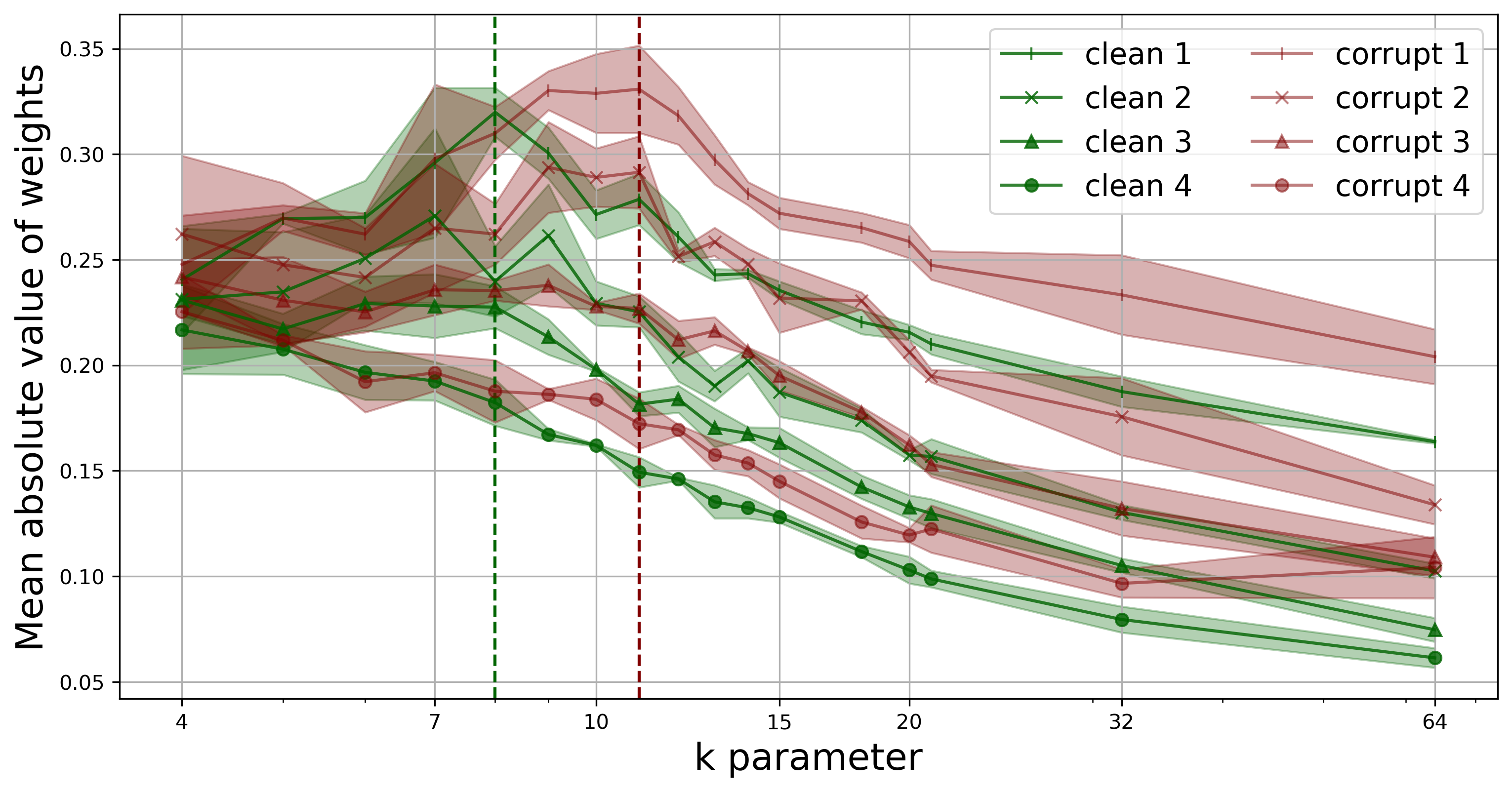}
    \includegraphics[width=0.49\linewidth]{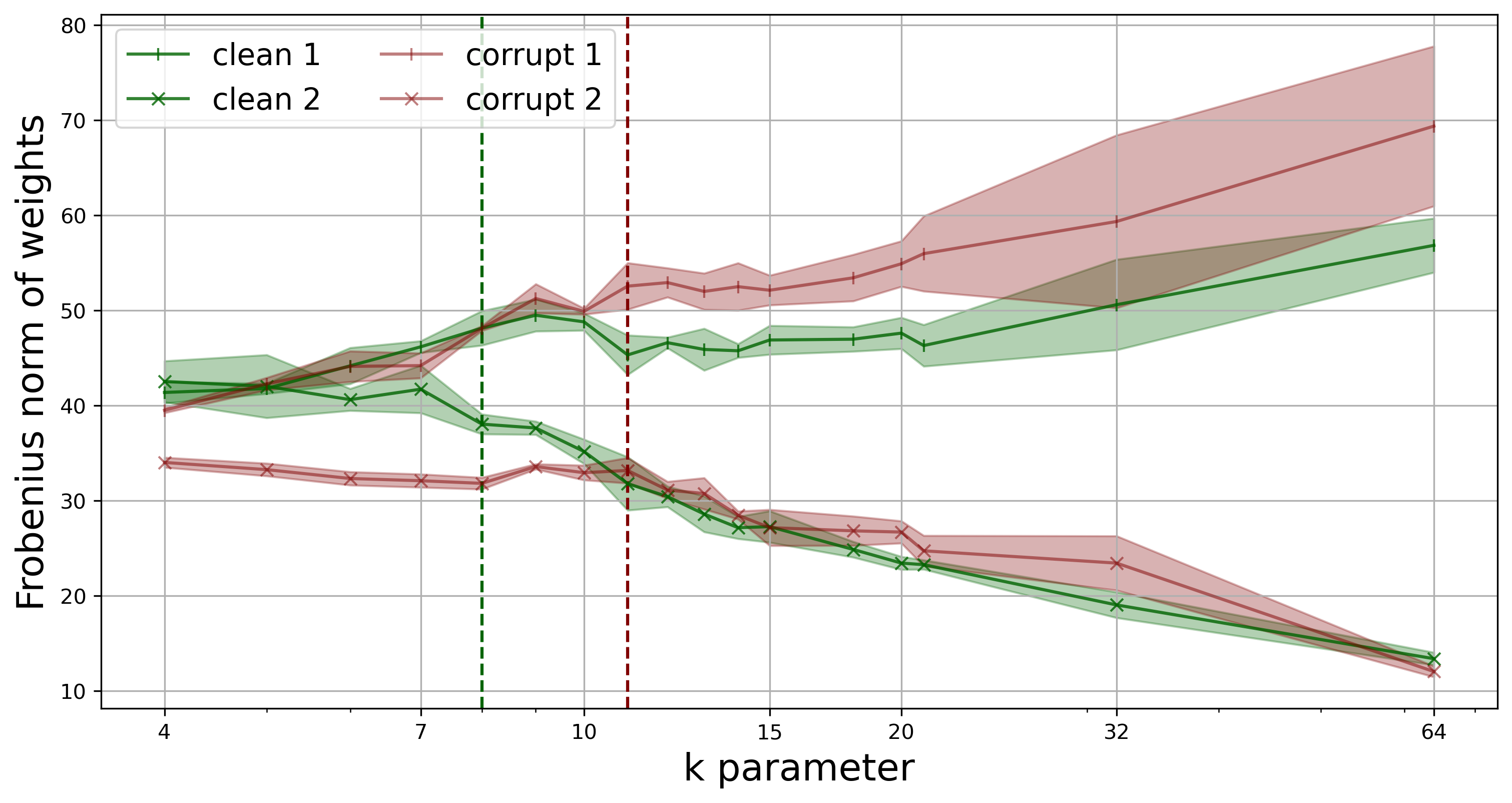}
    \includegraphics[width=0.49\linewidth]{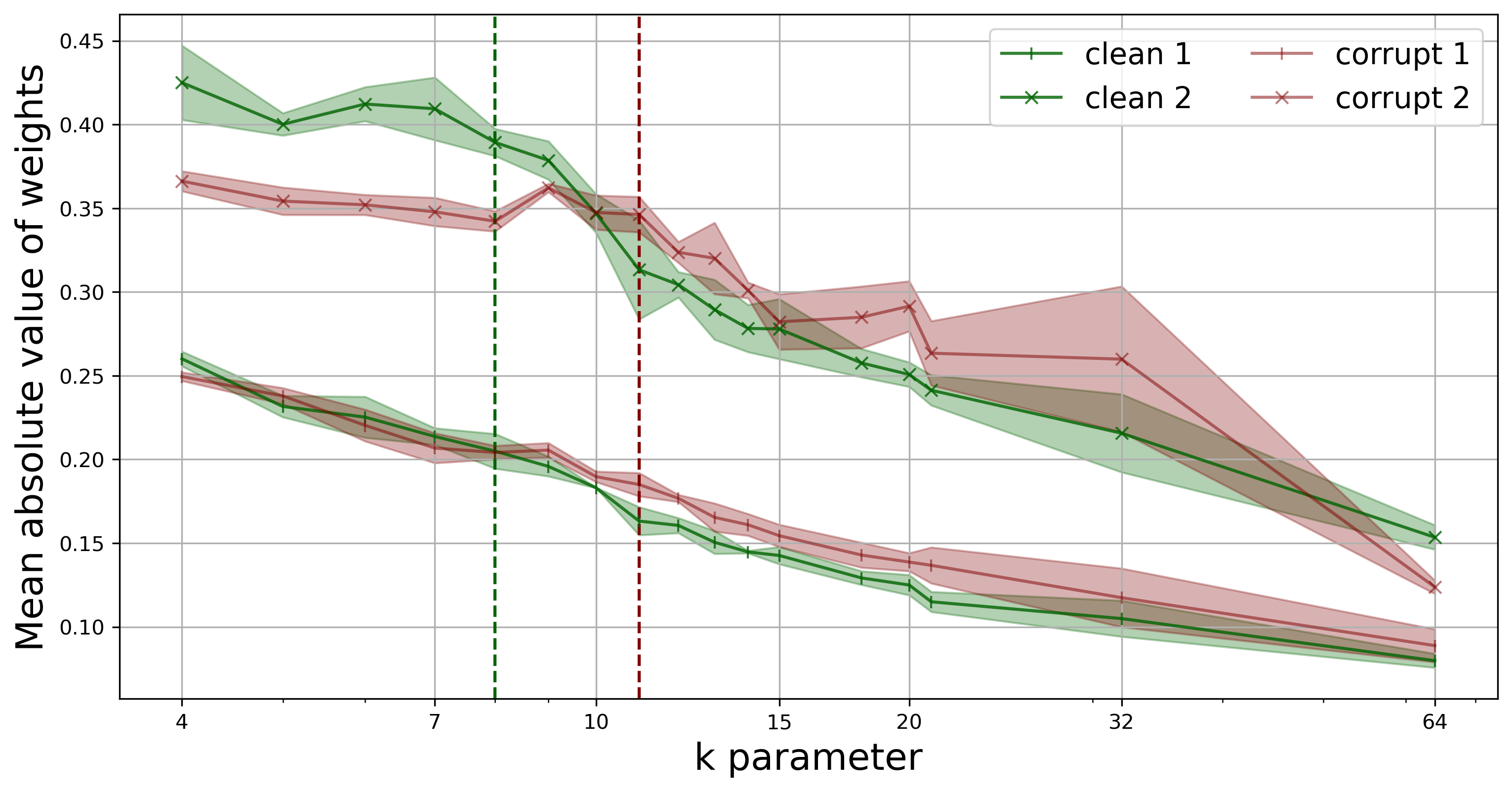}
    \caption{Parameter size per layer as model representational capacity is increased. Left: Frobenius norm. Right: Mean absolute value. Top: Convolutional layer weights. Bottom: Linear layer weights. 
    Vertical lines indicate point of maximum fragmentation.
    \label{fig:parameter_norms}} 
\end{figure}

Results are shown in Figure \ref{fig:parameter_norms}, with convolutional layer weights and linear layer weights displayed separately.  Lines and shaded regions indicate mean and standard deviations over seeds, respectively.
Dashed lines indicate the point of maximum fragmentation for the clean (green) and corrupted (red) models, respectively. 
(Compare Figure \ref{fig:input_and_hidden_dd_fragmentation}.)

We find that the Frobenius norms of the weights themselves, as well as the mean absolute values of the linear layer weights display highly regular behaviour as capacity is increased, with no link to points of fragmentation. 
On the other hand, for the {\em two shallowest} convolutional layers, the mean absolute values of the weights do show an  increase around the area of fragmentation. 
This occurs for both the clean and label-corrupted models, even though fragmentation occurs less in the first set than the second set.
We therefore conclude that while there is a potential link between the mean absolute weights of the shallower convolutional layers and fragmentation, this alone does not explain the phenomenon.

We also investigate the relationship between fragmentation and the distance a parameter moved during training (not shown here). Specifically, we measure the Frobenius norm of the difference between the parameter tensors before and after training. The parameter distance from initialization has been previously explored as a complexity measure~\cite{distance_from_init, fantastic_gen} and also served as a baseline in the PGDL challenge. 
We find that the distance a parameter is moved is highly correlated with the parameter norm itself: the precise value at initialization is overshadowed by the size of the parameter after training, and the difference between parameter norm and parameter movement as metric is negligible.



\end{document}